\documentclass[preprint,12pt,3p]{elsarticle}

\usepackage{lineno,hyperref}
\usepackage{amsmath,amssymb,amsfonts}
\usepackage{algorithmic}
\usepackage{algorithm}
\usepackage{graphicx}
\usepackage{textcomp}
\usepackage{subfigure}
\usepackage[misc]{ifsym}
\usepackage{color}
\usepackage{booktabs}
\usepackage{multirow}
\usepackage{times}
\usepackage{mathrsfs}
\journal{Journal of Visual Communication and Image Representation}

\begin{document}

\begin{frontmatter}

\title{Accurate Bounding-box Regression with Distance-IoU Loss for Visual Tracking}

\author[a,c]{Di Yuan\corref{cor1}}
\ead{dyuanhit@gmail.com}
\author[b]{Xiu Shu}
\author[c]{Nana Fan}
\author[d]{Xiaojun Chang}
\author[c,e]{Qiao Liu}
\author[c]{Zhenyu He}
\cortext[cor1]{Corresponding author.}
\address[a]{Guangzhou Institute of Technology, Xidian University, Guangzhou 510555, China}
\address[b]{School of Science, Harbin Institute of Technology, Shenzhen 518055, China}
\address[c]{School of Computer Science and Technology, Harbin Institute of Technology, Shenzhen 518055, China}
\address[d]{ School of Computing Technologies, RMIT University, VIC 3046, Australia}
\address[e]{National Center for Applied Mathematics, Chongqing Normal University, Chongqing, 401331, China}

\begin{abstract}
Most existing trackers are based on using a classifier and multi-scale estimation to estimate the target state.
Consequently, and as expected, trackers have become more stable while tracking accuracy has stagnated.
While trackers adopt a maximum overlap method based on an intersection-over-union (IoU) loss to mitigate this problem, there are defects in the IoU loss itself, that make it impossible to continue to optimize the objective function when a given bounding box is completely contained within/without another bounding box; this makes it very challenging to accurately estimate the target state.
Accordingly, in this paper, we address the above-mentioned problem by proposing a novel tracking method based on a distance-IoU (DIoU) loss, such that the proposed tracker consists of target estimation and target classification.
The target estimation part is trained to predict the DIoU score between the target ground-truth bounding-box and the estimated bounding-box.
The DIoU loss can maintain the advantage provided by the IoU loss while minimizing the distance between the center points of two bounding boxes, thereby making the target estimation more accurate.
Moreover, we introduce a classification part that is trained online and optimized with a Conjugate-Gradient-based strategy to guarantee real-time tracking speed.
Comprehensive experimental results demonstrate that the proposed method achieves competitive tracking accuracy when compared to state-of-the-art trackers while with a real-time tracking speed.
\end{abstract}

\end{frontmatter}

\section{Introduction}
Target tracking is a very hot and challenging visual task.
Trackers need to learn a target appearance model that relies on the given information of the target in the initial frame.
The learned model needs a strong generalization ability for the target appearance state.
The target tracking task in question could be divided into two parts: target classification and target estimation.
For target classification, it is a rough way of distinguishing the target from the background.
While the target estimation is used to accurately predict the target bounding box.

\begin{figure}[!t]
\centering
\includegraphics[width=0.95\textwidth]{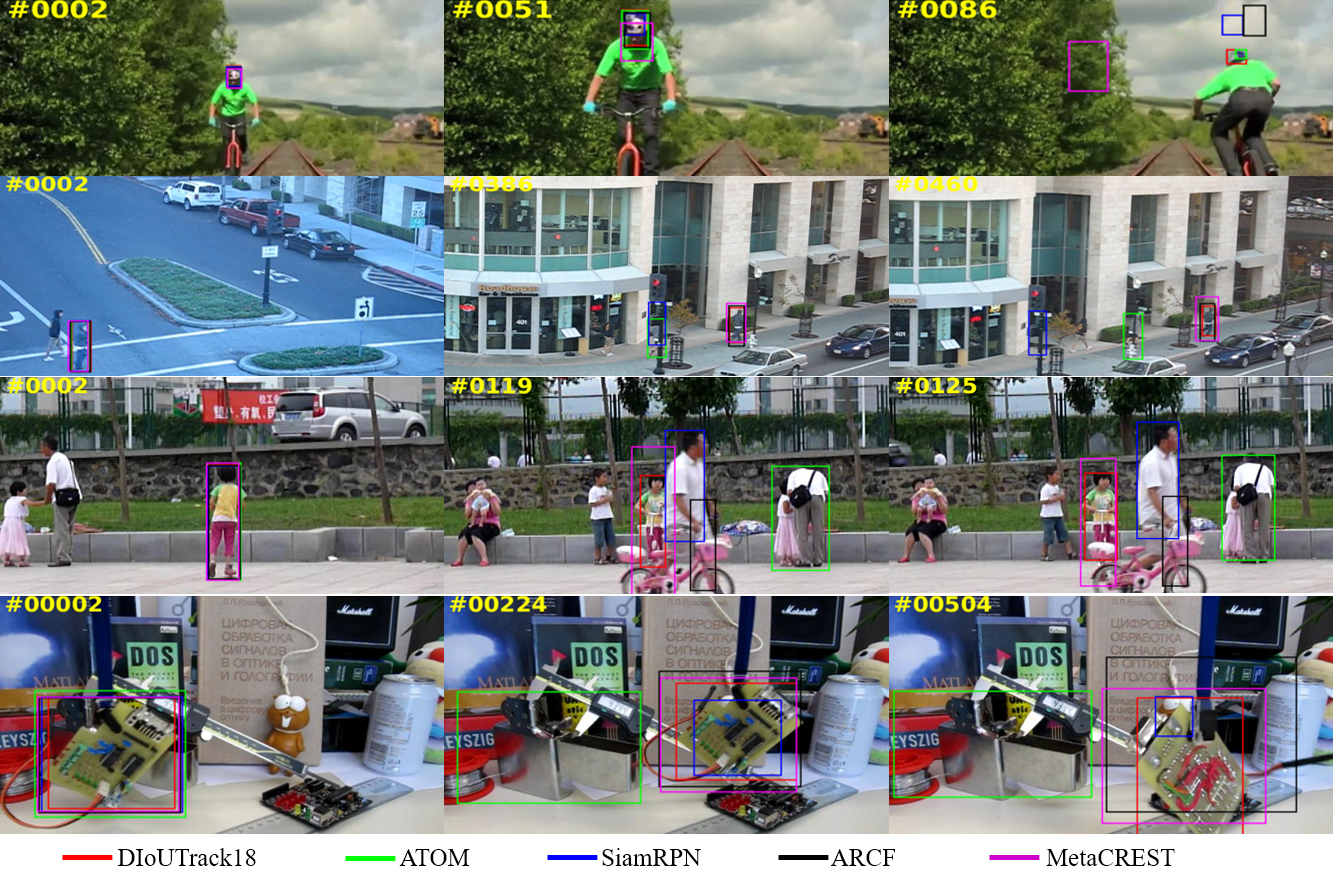}
\caption{A visual comparative experiment of our proposed tracker with other trackers.}
\label{fig1}
\end{figure}

Recently, target tracking research has tended to focus on the target classification component. Within these studies, most researchers focus on designing a robust classifier which is based on e.g. discriminative correlation filters \cite{GFSDCF,BACF,ARCF,ASTCA}, and exploiting some deep features \cite{MDNet,DaSiamRPN,TADT,TBC,DMLST} to achieve good tracking accuracy.
However, progress in the target estimation component has been slower than expected.
Most of the current representative trackers still adopt a multi-scale search strategy to estimate the target boundary box.
Such as, the MCPF \cite{MCPF} tracker handles the scale variation via a particle sampling strategy, while the ASRCF \cite{ASRCF} tracker only uses five-scale HOG features for scale estimation; moreover, the MetaCREST \cite{MetaSDNet} tracker extracts search patches in different scales to conduct target estimation.
Although the trackers mentioned above have achieved some good tracking performance, this multi-scale search strategy cannot accurately estimate the real target state.
By contrast, the SiamRPN \cite{SiamRPN} tracker adopts a bounding box regression method for target estimation, while the ATOM \cite{ATOM} tracker employs an overlap prediction network to estimate the target state; unfortunately, however, both of these trackers still struggle in cases of occlusion, deformation, etc. (shown in Figure \ref{fig1}).
Therefore, currently available target estimation methods cannot meet the requirements of practical applications.

\begin{figure}[!t]
\centering
\includegraphics[width=0.7\textwidth]{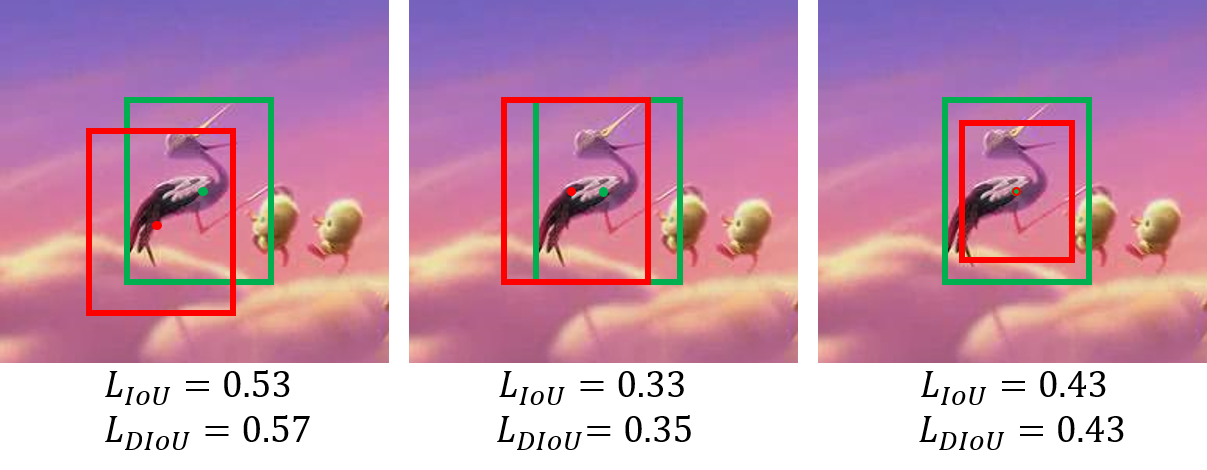}
\caption{Comparison of the DIoU loss and the IoU loss in some different cases. \textcolor{green}{Green} means ground-truth and \textcolor{red}{red} means predicted bounding box.}
\label{comparisonloss}
\end{figure}

Accordingly, in this work, our goal is to create a tracker that can improve tracking accuracy while also ensuring the tracker's robustness.
The tracking framework that we used include target estimation and classification.
Inspired by the DIoU \cite{DIoU} loss for bounding box regression in the object detection task, we learn the target estimation part so that it can predict the Distance Intersection over Union (DIoU) score between target ground-truth and an estimated bounding box.
In each tracking frame, the final target bounding-box is determined by maximizing the predicted DIoU score of some proposals and target bounding box in a reference frame.
It should be noted here that our DIoU score differs from the IoU score in some cases (shown in Figure \ref{comparisonloss}). Specifically, the loss of our DIoU-based network is higher than that of the IoU-based network when the centers of the two bounding-boxes do not coincide, which forces the two boundary boxes to quickly reach a state of the center overlaps.
In other words, it is easier for DIoU-based trackers to get accurate tracking results.

For the online target tracking phases, moreover, we choose a simple but effective two-layer fully convolutional network as the target classification part, due to it can provide high robustness in the complex tracking scenarios.
To ensure the real-time tracking speed, we follow the ATOM \cite{ATOM} tracker, which addresses the problem of efficient online optimization by employing a Conjugate-Gradient-based method.
The process of our online target tracking phases is simple: following model initialization, the target classification, target estimation, and model updating processes execute alternately until the entire tracking task is complete.
The main contributions are summarized as follows:
\begin{itemize}
\item We formulate a novel DIoU network-based bounding-box regression model for target tracking. While preserving the advantages offered by the IoU network in tracking tasks, the DIoU network can be deployed to directly minimize the distance between the ground-truth bounding box and predicted bounding box, an approach that allows the tracker to obtain more accurate tracking results.
\item We adopt a Conjugate-Gradient-based strategy to ensure that the optimization problem in the target classification component can be addressed efficiently online.
\item Extensive experiments have verified that our tracking method is more competitive than other state-of-the-art trackers on seven challenging datasets: OTB100 \cite{OTB100}, UAV123 \cite{UAV123}, TrackingNet \cite{TrackingNet}, LaSOT \cite{Lasot}, GOT10k \cite{GOT10k}, VOT2018 \cite{VOT2018} and VOT2019 \cite{VOT2019}.
\end{itemize}

\section{Related Work} \label{Rw}
At present, most target trackers either under the detection-based framework and or under the template matching-based framework.
Trackers based on the detection framework treat the target tracking task as a classification problem and distinguish the target from the background by modeling the target appearance.
While, trackers based on the template matching framework typically use a Siamese network to determine the target location utilizing spatial cross-correlation, which can be used to the most relevant candidates for the target.

\subsection{Tracking-by-detection frameworks}
There are many tracking approaches that combine tracking and detection in some respect \cite{BACF,TIOIF,TLD,GBBMT,ALT,BSegm,MTRE,SNNL}. In \cite{TLD}, the TLD tracking framework divides tracking task as tracking, learning and detection sub-tasks. Each of these three parts complements each other to enable the target tracking task to be completed.
In \cite{TIOIF}, Wang $et$ $al.$ demonstrate  that tracking different objects could be formulated as a network-flow mixed-integer program. Lan $et$ $al.$ \cite{MTRE}  propose to the target tracker in a frame-by-frame manner by exploring the time, space, and multi-camera relationship of detection hypotheses shortly frames.
Other trackers have integrated the detector within a particle filter tracking framework \cite{MCPF,PFCFT}.
Among these detection-based tracking methods, DCF-based tracking methods achieved some promising performance \cite{BACF,KCF,STRCF,TRBACF}. These DCF-based tracking methods learn a correlation filter from target ground-truth provided in the initial image frame to discriminate between target and background. In \cite{KCF}, Henriques $et$ $al.$ derive a kernelized correlation filter with the exact same complexity as its linear counterpart, while also proposing a fast multi-channel correlation filter; this allows the KCF tracker to achieve promising tracking accuracy and fast-tracking speed compared to other trackers of the same period.
However, DCF-based trackers can not model the background well. To resolve this issue, Kiani $et$ $al.$ \cite{BACF} proposes a background-aware correlation filter-based tracker to model both target and background. By introducing a temporal regularizer to the DCF-based trackers, it has been able to achieve a competitive tracking result\cite{ASRCF,STRCF}. To improve tracking accuracy, a group feature selection strategy has been proposed under the DCF-based tracking framework that can select group features across channels and spatial dimensions to determine the structural correlation between feature channel and filter system \cite{GFSDCF}.
The DCF-based trackers mentioned above are only able to determine the target center location, most of these trackers use a multi-scale search strategy to predict the target state, which usually results in relatively inaccurate tracking results\cite{TFCR,ASCT}.
The recently proposed ATOM \cite{ATOM} tracker incorporates IoU modulation and IoU prediction to improve tracking performance.
However, the IoU loss has an inherent defect: that is, when one bounding-box is completely inside the other, the IoU loss does not change; the centers of these two bounding-boxes do not necessarily overlap \cite{ATCC}. Accurate target boundary box positioning is very important for tracking tasks, meaning that further improvement of the IoU-based trackers is required.

\subsection{Template matching frameworks}
Template matching-based tracking frameworks typically use a Siamese network as the similarity measurement network \cite{SINT,SiamFC,SRST,SiamRPN,SSDCT,DCFHT}. As the first Siamese network-based tracker, SINT \cite{SINT} simply matches the initial target with proposals selected in the current frame and given the most similar proposal as the tracking target.  Despite its simple network structure, the SINT tracker achieves efficient tracking performance, but suffers from a very slow tracking speed.
In \cite{SiamFC}, the SiamFC tracker was proposed with the aim of achieving a high tracking accuracy and a fast-tracking speed. In response to this work, there are many trackers that extend the SiamFC architecture for the target tracking task \cite{SiamRPN,VOTAS,SiamDW,SASiam,SiamRPN++}.
The SiamRPN \cite{SiamRPN} tracker joins the RPN network under the Siamese-based tracking framework. As a result of the region proposal refinement, the whole tracking process is simplified without affecting tracking performance. Both the DaSiamRPN \cite{DaSiamRPN} tracker and the SiamRPN++ \cite{SiamRPN++} tracker, as improved versions of the SiamRPN \cite{SiamRPN} tracker, improve the tracking performance in different ways.
Although Siamese-based trackers provide an acceptable balance between tracking speed and accuracy. Most of the Siamese-based trackers are difficult to classify targets effectively due to a lack of online model updating.
Unlike these trackers, our proposed tracker not only has an offline training of the model but also offers a model update strategy during the online tracking phase, which allows for accurately estimated the target state when the target appearance changes dramatically.

\subsection{Bounding box regression for tracking task}
In target tracking tasks, a rectangular bounding box is usually utilized to display the target location. Accurate target boundary box estimation is a complex task, which depends strongly on the target location and scale. The target location is the key to determine the bounding box center.  While the target scale is the key to whether the bounding box can accurately return to the state of the target or not. As a result, many trackers use lots of off-line training to try to get enough priors \cite{SiamRPN,DaSiamRPN}.
Notably, the DaSiamRPN \cite{DaSiamRPN} tracker has obtained sufficient prior knowledge based on the off-line training, and therefore obtained promising results on bounding box regression.
However, these trackers are always affected when they encounter the target classification problem.
Different from the Siamese-based tracking methods, the ATOM \cite{ATOM} tracker and some of its variants \cite{DiMP,PrDiMP} trains a target estimation strategy to calculate the IoU overlap scores of proposals and the reference target. By maximizing the IoU overlap score, the ATOM \cite{ATOM} tracker can predict a compact bounding-box of the tracking target. GIoU \cite{GIoU} loss has also been proposed to tackle the gradient vanishing issues, but is affected by slow convergence and inaccurate regression.
In comparison, the DIoU \cite{DIoU} loss offers faster convergence and better bounding box regression accuracy. Accordingly, we utilize the DIoU loss to improve the IoU-based tracker to achieve some competitive tracking results.

\section{Proposed Method} \label{OurM}
\begin{figure}[!t]
\centering
\includegraphics[width=0.85\textwidth]{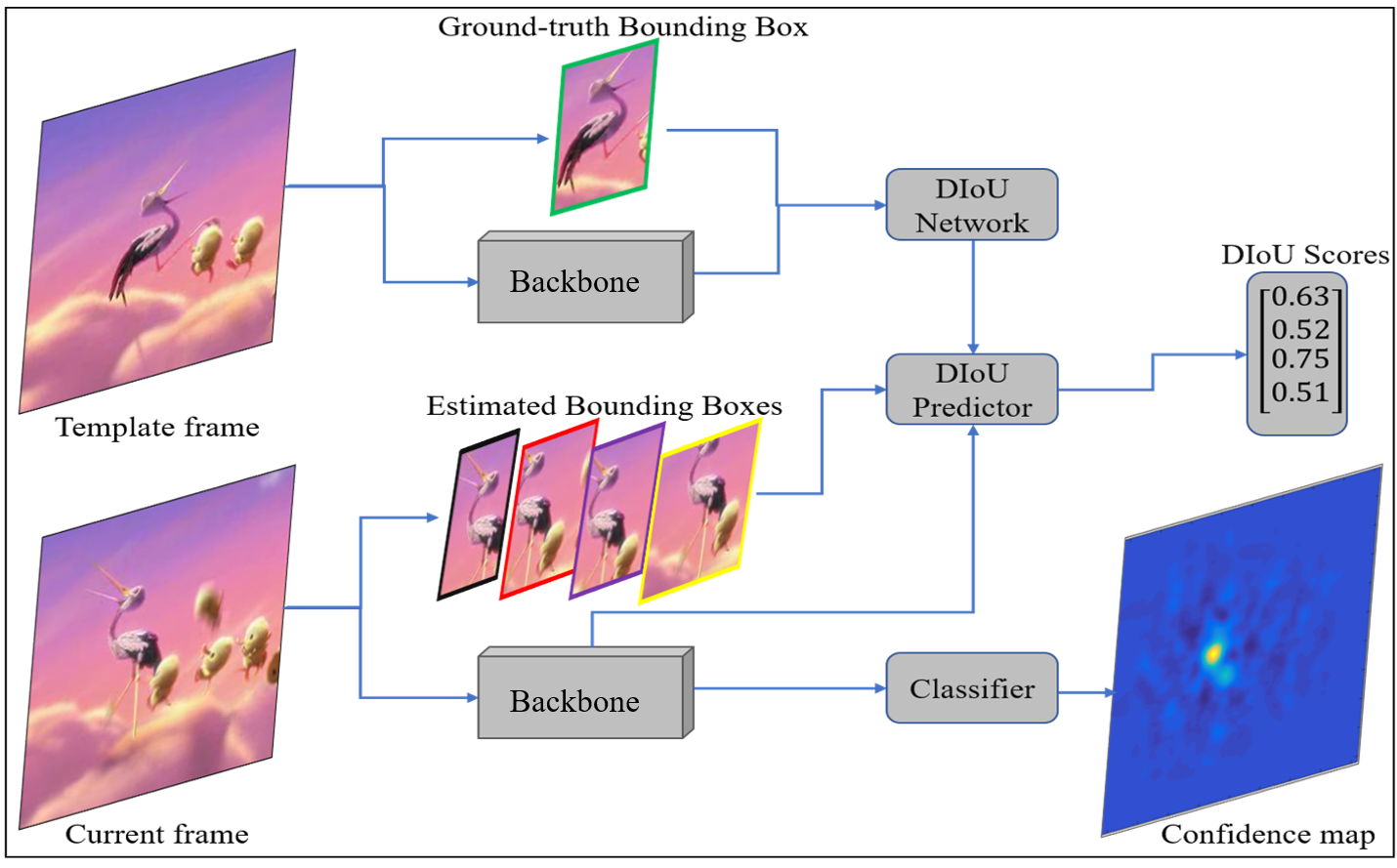}
\caption{Architecture of the proposed method for target tracking task.
The DIoU predictor is pre-trained on some large training sets to predict the DIoU score of the target candidates. The target classifier is trained online to output the corresponding confidence map.}
\label{Overview}
\end{figure}

We follow the process used in ATOM \cite{ATOM} and divide the tracker into two components: an offline learned target estimation component and an online learned target classification component. In other words, we separate the tracking problem into two sub-problems (classification and estimation). The whole tracking architecture is shown in Figure \ref{Overview}.

\subsection{Target estimation via bounding box regression} \label{TEBBR}
As described in the ATOM \cite{ATOM} tracker, the target state estimation aims to accurately predict the target bounding box by means of a rough initial estimate.
The ATOM tracker uses an improved IoUNet \cite{IoUNet} for the target estimation; this means that given image ($x$) and bounding box estimate of a target ($B$), the IoUNet can calculate IoU score between estimate bounding box ($B$) and target ground-truth ($B^{gt}$).
\begin{equation}
 IoU =\frac{B \cap B^{gt}}{B \cup B^{gt}}.
\end{equation}

The prediction network to pool the region in the image $x$ given by the estimate bounding-box, resulting in a determined size feature map.
The ROI Pooling is differentiable and can be used to improve the predicted bounding box by maximizing IoU score.
However, the IoU-based bounding-box regression for target tracking has an obvious drawback: when one bounding-box is located entirely within another bounding-box, the objective function based on the IoU loss is no longer optimized (see the right sub-figure of Figure \ref{Dloss}). However, the prediction bounding-box may not be optimal; in other words, the tracking results are not accurate.
We, therefore, propose an improved IoU loss-based bounding box regression method to ensure the tracking accuracy.

\subsection{Bounding-box regression by DIoU loss} \label{DIoUloss}
\begin{figure}[!t]
\centering
\includegraphics[width=0.5\textwidth]{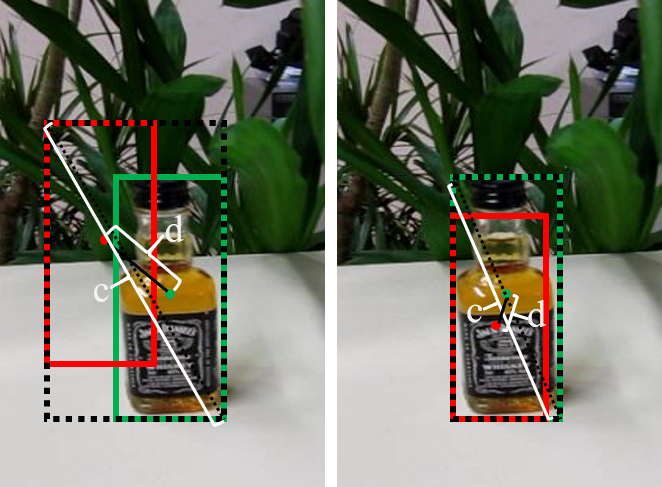}
\caption{DIoU loss for bounding box regression. $d=\rho(b,b^{gt})$ is the distance of central points of these two boxes.}
\label{Dloss}
\end{figure}

We take inspiration from the DIoU \cite{DIoU}, a method that was recently proposed for object detection task, as this results in much faster convergence in training than the IoU loss.
The loss function based on IoU can be defined according to the following format:
\begin{equation}
L = 1 - IoU + P(B,B^{gt}),
\end{equation}
where $P(B,B^{gt})$ is a penalty term.
When the penalty term $P(B,B^{gt}) = 0$, the loss function will degenerate into the IoU loss.
The DIoU score could be calculated as follows:
\begin{equation} \label{DIoUs1}
 S_{DIoU} = IoU - \lambda \frac{\rho^2(b,b^{gt})}{c^2},
\end{equation}
where $b$ and $b^{gt}$ are the central points of $B$ and $B^{gt}$, $c$ is the diagonal length of the minimized enclosing bounding box $C$ that covers $B$ and $B^{gt}$ (see Figure \ref{Dloss}), and $\lambda$ is a parameter to balance IoU score and penalty term.
In general, the DIoU score is always lower than the IoU score, and they are equal if and only if the centers of the two bounding boxes overlap.
This also brings the prediction bounding box depended on the DIoU score is closer to the reference bounding box center.
The DIoU loss could be defined as follows:
\begin{equation} \label{DIoUeq1}
L_{DIoU} = 1 - IoU + \lambda \frac{\rho^2(b,b^{gt})}{c^2},
\end{equation}
where $\rho(.)$ is Euclidean distance.
The DIoU score can directly reflect the overlap degree between $B$ and $B^{gt}$, as well as whether the center position of these two bounding boxes is the same.
The penalty term $\lambda \frac{\rho^2(b,b^{gt})}{c^2}$ directly minimizes the distance between the central points of these two bounding boxes.
When $\lambda=0$, the DIoU loss will degenerate to the IoU loss. In addition, the value of $\lambda$ only affects the training speed of the model and has no obvious influence on the tracking performance of the trained model. Therefore, we set $\lambda=1$ in this paper.
The DIoU-based network is trained by minimizing DIoU losses between candidate samples and reference targets.
The target boundary box is predicted by maximizing the DIoU prediction overlap score.

\subsection{Target classification for tracking}
Although the target estimation component can provide an accurate bounding-box for the tracking task, it cannot make robust distinctions between the target and the background.
In this section, we introduce a robust target classifier that can accurately determine the target and background, regardless of whether or not the tracking scene is disturbed.
Different from target estimation, target classification component can be trained directly online and used for target confidence score prediction.
Refer to literature \cite{ATOM}, the target classification component we used can be defined as follows:
\begin{equation} \label{clssmodel}
 f(z;w) = \phi_2(w_2 * \phi_1(w_1 * z)),
\end{equation}
where $z$ denotes the feature map of the target, while $w = \{ w_1,w_2 \}$ are parameters, $\phi_1, \phi_2$ are the activation functions in the network.

In order to achieve a fast tracking speed, we refer to the DCF-based trackers \cite{BACF,STRCF} to build a $l_2$ error-based model, as follows:
\begin{equation}\label{eq:dcf1}
L(w) = \lVert f(z;w) - y \rVert^2 + \zeta \lVert w \rVert^2,
\end{equation}
where $z$ is the training sample feature map and $y$represents the corresponding label with a Gaussian shape.
Generally, Eq. (\ref{eq:dcf1}) is optimized by stochastic gradient descent, which makes the tracking speed slow.
Similar with literature \cite{ATOM}, the object function (\ref{eq:dcf1})  can be formulated as a squared $L^2$ norm of the residual vector $L(w) = \lVert r(w) \rVert^2$. According to the first order Taylor expansion method, we can know that: $r(w + \Delta w) \approx r(w) + \frac{\partial r}{\partial w} \Delta w$. Using the quadratic
Gauss-Newton approximation, we can obtain:
\begin{equation}\label{eq:gn1}
L_w(\Delta w) =  \Delta w^T \frac{\partial r}{\partial w}^T \frac{\partial r}{\partial w} \Delta w + 2 \Delta w^T \frac{\partial r}{\partial w}^T r(w) + r(w)^T r(w),
\end{equation}
where the $\Delta w$ is a increment in the parameters $w$.
The Gauss-Newton problem (\ref{eq:gn1}) forms a positive definite quadratic function, it allows the Conjugate-Gradient method to be used to solve this problem. The Conjugate-Gradient method consists of simple vector operations, which can be implemented easily in code.
The most important part of Conjugate-Gradient is to find the optimal search direction $p$ and step size $\alpha$ in each iteration.
The search direction $p$ is determined by $\frac{\partial r}{\partial w}^T \frac{\partial r}{\partial w}p$.
For evaluate $\frac{\partial r}{\partial w}^T \frac{\partial r}{\partial w}p$, a vector $u$ with the same size of $r(w)$ has been considered, and $\frac{\partial }{\partial w}(r(w))^Tu = \frac{\partial r}{\partial w}^T u$ is the standard operation of the back-propagation procedure.
The Jacobian of the function $u \mapsto \frac{\partial r}{\partial w}^T u$  is trivially $\frac{\partial r}{\partial w}^T$, since the function is linear.
Each Conjugate-Gradient iteration requires two back-propagation: $q_1 = \frac{\partial r}{\partial w}p$ and $q_2 =  \frac{\partial r}{\partial w}^T q_1$.
More details can be found in literature \cite{ATOM}.

\subsection{Offline training for DIoU-based predictor}
The proposed DIoU prediction network is pre-trained offline by using labeled training images as in Eq. (\ref{DIoUeq1}).
Similar to \cite{ATOM}, we used the LaSOT dataset \cite{Lasot}, the TrackingNet \cite{TrackingNet} dataset and the COCO \cite{COCO} dataset as training data.
Each training image pair contains one template image and one test image.
For the template image, an image patch centered at the target has been cropped as a template sample; the template sample's size is $5$ times the length and width of the target size.
For the test image, we crop a similar image patch and add perturbations to simulate a real tracking scene.
The cropped image patches are resized in the same size to train the network.
We fixed all weights of our backbone network in the training phase and use L2 to train the DIoU-based predictor. The predictor was trained for $60$ epochs and batch size set to $64$.
We also utilize ADAM optimizer with an initial learning rate $lr = 10^{-3}$ and a decay factor $df = 0.2$ for every $15$ epochs.

\section{Experiments} \label{Exp}
Our experiments are performed in Python using PyTorch, and the tracking speed is over $50/40$ $fps$ with the backbone network ResNet18/ResNet50 on an NVIDIA GTX 2080Ti GPU.
To evaluate the tracking performance of the proposed tracking method, we make some experimental comparisons of our tracker with several state-of-the-art trackers on $7$ challenge datasets: OTB100 \cite{OTB100}, UAV123 \cite{UAV123},  TrackingNet \cite{TrackingNet}, LaSOT \cite{Lasot}, GOT10k \cite{GOT10k}, VOT2018 \cite{VOT2018} and VOT2019 \cite{VOT2019}.

\subsection{Tracking process} \label{DIoUTrack}
Once the DIoU estimates have been trained offline, the online tracking process of the proposed tracking method can be easily subdivided into the following four steps: model initialization, target classification, target estimation, and model update.

\noindent \textbf{Model Initialization.}
We use the ResNet as our backbone network to extract features.
Beginning with the initial target state, an image patch $5$ times the size of the target was cropped and extract features from patch size $288\times 288$ from the cropped patch.

\noindent \textbf{Target Classification.}
Following the ATOM \cite{ATOM} tracker, the target classification network in our tracker consists of a 2-layer CNN network.
The first layer consists of a $1 \times 1$ convolutional layer ($w_1$), while the second layer adopts a $4 \times 4$ kernel ($w_2$) with a single output channel. Where $\phi_1(t) = t, t \geq 0$ is an identity transformation and $\phi_2(t) = \alpha (e^{t/\alpha}-1), t \leq 0$ ($\alpha =0.05$ in this paper). Moreover, $\phi_2$ offers continuous differentiability and is thus good for optimization.
In the first frame, we generate $30$ training samples through data expansion, and optimize the parameters $w_1$ layer with $6$ rounds of Gauss-Newton iterations and $10$ rounds of Conjugate-Gradient iterations.
We then only optimize the $w_2$ layer with $1$ round of Gauss-Newton iterations and $5$ rounds of Conjugate-Gradient iterations for each 10th frame.

\noindent \textbf{Target Estimation.}
At current $t$-th frame, the position with the highest confidence score can be found by using the classification model (Eq. (\ref{clssmodel})).
After that, we can use this position as the target center point and generate $10$ bounding boxes randomly. The DIoU score of each bounding box was maximized by the offline trained target estimation network. The final state of the predicted target in the current frame is determined by the average of these bounding-boxes with top-k DIoU scores.

\noindent \textbf{Model Update.}
In the target classification phases, we adopt the $l_2$ classification error in the DCF-based tracking framework so that we can distinguish target from background. And we adopt a linear update strategy: $w = (1 - \delta)w_{t-1} + \delta w_t$ to update $w$, where $\delta$ is a learning rate.

\subsection{Ablation study}
We first give an ablation study on the LaSOT \cite{Lasot} and OTB100 \cite{OTB100} datasets to verify the effectiveness of each component in the proposed tracker. The backbone network we used in this part is ResNet18.
We mainly analyze the impact of the two main components (DIoU loss \& Conjugate-Gradient) in our tracker on tracking performance.
The experimental results are shown in Table \ref{table:ab}.
To avoid confusion, we state that trackers without the DIoU loss mean that they adopt the IoU loss to train their model, and trackers without the Conjugate-Gradient strategy mean that they only adopt the Gauss-Newton strategy for the model optimizing.
From this table we can know that the tracking performance of the tracker with the DIoU loss is significantly improved than the tracker without the DIoU loss, especially in the success score, it has about $11\%$ improvement on the LaSOT \cite{Lasot} dataset.
In addition, Table \ref{table:ab} also clearly reflects the tracking performance of the tracker with the Conjugate-Gradient strategy is better than the tracker without it.

\begin{table*} 
\renewcommand\arraystretch{1.2}
\renewcommand\tabcolsep{3pt}
\scriptsize
\centering
\caption{Comparison results of ablation study on the LaSOT \cite{Lasot} and OTB100 \cite{OTB100} datasets.}
\label{table:ab}
\begin{tabular}{c|c|c|c|c|c|c}
\toprule
\multirow{2}{*}{DIoU Loss} &  \multirow{2}{*}{Conjugate-Gradient} &\multicolumn{3}{c|}{ LaSOT} &\multicolumn{2}{c}{ OTB100}\\
\cline{3-5} \cline{6-7}
& & Precision scores & Normalized Precision scores & Success scores & Precision scores  & Success scores\\
\midrule
$\times$      & $\times$      & $35.0$ & $35.1$ & $36.6$  & $61.1$ & $47.0$ \\
$\times$      & $\checkmark$  & $44.5$ & $45.7$ & $42.9$  & $77.2$ & $58.7$ \\
$\checkmark$  & $\times$      & $42.1$ & $43.5$ & $46.3$  & $74.7$ & $61.6$ \\
$\checkmark$  & $\checkmark$  & $51.6$ & $54.7$ & $54.7$  & $89.0$ & $68.1$ \\
\bottomrule
\end{tabular}
\end{table*}

\subsection{State-of-the-art comparison}
We present some quantitive comparisons of our DIoUTrack with a number of state-of-the-art trackers on the $7$ most challenging single target tracking datasets.
Since we use two backbone networks (ResNet18 / ResNet50), we give the tracking results of the corresponding trackers (DIoUTrack18 / DIoUTrack50).

\noindent \textbf{Experiment on OTB100 \cite{OTB100} dataset}: The OTB100 dataset includes $100$ testing sequences and the tracking accuracy of each tracker is evaluated by precision (a center position distance between the predicted and ground-truth of the target that is $\leq$ a fixed threshold (such as $20$ pixel values) is considered to have successfully tracked the target) and success (an area-under-curve (AUC) $\geq$ $0.5$ is considered to have successfully tracked the target).
We draw some experimental comparisons of the proposed DIoUTrack and several state-of-the-art trackers (namely ATOM \cite{ATOM}, GradNet \cite{GradNet}, GCT\cite{GCT}, ARCF \cite{ARCF}, UDT \cite{UDT}, MetaCREST \cite{MetaSDNet}, SiamRPN \cite{SiamRPN}, SiamRPN++ \cite{SiamRPN++}, PTAV \cite{PTAV}, DiMP18 and DiMP50 \cite{DiMP}) on this dataset.
Table \ref{tab:OTB100} presents the results of these comparisons over all $100$ testing videos. From this table, we can know the proposed DIoUTrack50 achieved the best tracking accuracy in both precision and success index.
The SiamRPN \cite{SiamRPN} tracker employs a bounding-box regression strategy, while the ATOM \cite{ATOM} tracker adopts an improved bounding-box regression model based on the IoUNet to estimate the target state.
Compared to other trackers, the ATOM \cite{ATOM} tracker achieves the acceptable success score and precision score ( $66.1\%$ / $86.2\%$), while the DiMP18 \cite{DiMP} tracker achieves good tracking accuracy ($66.0\%$ / $87.8\%$); however, our DIoUTrack18 with the same backbone network (ResNet18), due to employing a DIoU network-based bounding-box regression model for target estimation, significantly outperforms the ATOM tracker and the DiMP18 tracker by achieving a success score of $68.1\%$ and a precision score of $89.0\%$.

\begin{table*}[http]
\renewcommand\arraystretch{1.2}
\renewcommand\tabcolsep{2pt}
\scriptsize
\centering
\caption{Comparison results on OTB100 dataset.
The top-3 scores are highlighted in {\color{red}\text{red}}, {\color{blue}\text{blue}} and {\color{green}\text{green}}, respectively.}
\label{tab:OTB100}
\begin{tabular}{lccccccccccccc}
\toprule
\multirow{2}{*}{Trackers} & DIoUTrack18 & DIoUTrack50 & DiMP18 & DiMP50 & ATOM & GradNet & GCT & ARCF & UDT & MetaCREST & SiamRPN  & PTAV & SiamRPN++\\
& Ours & Ours &\cite{DiMP} &\cite{DiMP}  &\cite{ATOM} &\cite{GradNet} &\cite{GCT} & \cite{ARCF} &\cite{UDT} &\cite{MetaSDNet} &\cite{SiamRPN}  &\cite{PTAV} &\cite{SiamRPN++}\\
\midrule
Precision & $89.0$  & \color{red}\text{92.3} & 87.8 &  \color{green}\text{89.9} & 86.2 & 86.1 & 85.9 & 81.8 & 76.0 & 85.7  & 85.1 & 84.8 & $\color{blue}\text{91.6}$ \\
Success  & $68.1$ &$\color{red}\text{71.0}$ & 66.0 &\color{green}\text{68.7} & 66.1 & 63.9 & 64.8 & 61.7 & 59.4 & 63.7 & 63.7 & 63.4 & $\color{blue}\text{69.6}$\\
\bottomrule
\end{tabular}
\end{table*}

\noindent \textbf{Experiment on UAV123 \cite{UAV123} dataset}: This UAV123 dataset consists of 123 testing aerial video sequences, and the performance is evaluated in the same way as the OTB100 dataset. To evaluate the tracking performance of the proposed DIoUTrack, we report some experimental comparisons of our tracker and several other state-of-the-art trackers (namely ATOM \cite{ATOM}, GFSDCF \cite{GFSDCF}, LDES \cite{LDES}, UDT \cite{UDT}, STRCF \cite{STRCF}, ARCF \cite{ARCF}, GCT \cite{GCT}, SiamRPN++ \cite{SiamRPN++}, SiamRPN \cite{SiamRPN}, DaSiamRPN \cite{DaSiamRPN}, DiMP18 and DiMP50 \cite{DiMP}) on this dataset.
Table \ref{tab:UAV123} presents the precision and success scores on $123$ video sequences.
DaSiamRPN \cite{DaSiamRPN}, SiamRPN++ \cite{SiamRPN++} and their predecessor SiamRPN \cite{SiamRPN} adopt a bounding-box regression-based target estimation component.
Compared to other tracking methods, DiMP50 \cite{DiMP} achieves superior tracking performance in terms of AUC (65.4\%) and precision (85.8\%) indexes.
While, SiamRPN++ \cite{SiamRPN++} achieves good tracking performance in terms of AUC (61.3\%) and precision (80.7\%) indexes.
However, the proposed DIoUTrack50 with the same backbone network (ResNet50), which employs a distance-IoU network-based bounding-box regression model for target estimation,  outperforms the DiMP50 \cite{DiMP} tracker and the SiamRPN++ \cite{SiamRPN++} tracker, achieving an AUC of 65.5\% and a precision of 86.6\%.
Compared to the ARCF \cite{ARCF}, which is a tracker specifically designed to track targets in a drone scenario. Our DIoUTrack achieves an improvement of more than 15\% in each index.

\begin{table*}[http]
\renewcommand\arraystretch{1.2}
\renewcommand\tabcolsep{1pt}
\scriptsize
\centering
\caption{Comparison results on UAV123 dataset. The top-3 scores are highlighted in {\color{red}\text{red}}, {\color{blue}\text{blue}} and {\color{green}\text{green}}, respectively.}
\label{tab:UAV123}
\begin{tabular}{lcccccccccccccc}
\toprule
\multirow{2}{*}{Trackers}  & DIoUTrack18 & DIoUTrack50 & DiMP18 & DiMP50 & ATOM & GFSDCF & LDES & UDT & STRCF & ARCF & GCT & SiamRPN++ & SiamRPN & DaSiamRPN \\
& Ours & Ours &\cite{DiMP} &\cite{DiMP} &\cite{ATOM} & \cite{GFSDCF} & \cite{LDES} & \cite{UDT} & \cite{STRCF} & \cite{ARCF} & \cite{GCT} & \cite{SiamRPN++} & \cite{SiamRPN} & \cite{DaSiamRPN}\\
\midrule
Precision & $\color{green}\text{85.4}$ & \color{red}\text{86.6} & 83.0 & \color{blue}\text{85.8} & 84.4 & 76.7 & 70.0 & 66.7 & 67.8 & 67.6 & 73.2  & 80.7 & 79.6 & 74.8\\
Success  & $\color{green}\text{64.3}$  & \color{red}\text{65.5} & $\color{green}\text{64.3}$  & \color{blue}\text{65.4} & 63.2 & 53.4 & 49.2 & 47.9 & 47.7 & 47.0 & 50.8 & 61.3 & 58.6 & 52.7\\
\bottomrule
\end{tabular}
\end{table*}

\noindent \textbf{Experiment on TrackingNet \cite{TrackingNet} dataset}: TrackingNet is containing a test set of $511$ video sequences.
To verify the tracking results of our DIoUTrack, we make some comparisons of its performance on the TrackingNet test set with several state-of-the-art trackers, namely ATOM \cite{ATOM}, SPM \cite{SPM}, GFSDCF \cite{GFSDCF}, C-RPN \cite{C-RPN}, UpdateNet \cite{UpdateNet}, DiMP18 \cite{DiMP}, DiMP50 \cite{DiMP}, UPDT \cite{UPDT}, ECO \cite{ECO}, SiamRPN++ \cite{SiamRPN++} and DaSiamRPN \cite{DaSiamRPN}. Table \ref{tab:TrackingNet} presents the comparison results in precision score, normalized precision score, and success score. From this table, it is evident that our DIoUTrack50 achieves the best scores in these three metrics. In terms of precision, our DIoUTrack50 outperforms the second-best tracker, DiMP50 \cite{DiMP}, by 1.3\%; moreover, compared to the Siamese framework-based DaSiamRPN \cite{DaSiamRPN} tracker, the proposed DIoUTrack50 achieves a greater than 18\% improvement in success and an improvement of over 28\% in precision. Finally, compared with the IoUNet-based ATOM \cite{ATOM} tracker, our DIoUTrack18 with the same backbone network (ResNet18) achieves an improvement of more than 2\% on each index. All of these comparative results show that the adopted Distance-IoU loss can effectively improve the target bounding-box regression model for accurate target location and estimation.

\begin{table*}
\renewcommand\arraystretch{1.2}
\renewcommand\tabcolsep{1pt}
\scriptsize
\centering
\caption{Comparison results on TrackingNet test set. The top-3 scores are highlighted in {\color{red}\text{red}}, {\color{blue}\text{blue}} and {\color{green}\text{green}}, respectively.}
\label{tab:TrackingNet}
\begin{tabular}{lccccccccccccc}
\toprule
\multirow{2}{*}{Trackers} & DIoUTrack18 & DIoUTrack50 & ATOM & SPM & GFSDCF & C-RPN & UpdateNet & DiMP18  & DiMP50 & UPDT & ECO & SiamRPN++ & DaSiamRPN\\
& Ours & Ours & \cite{ATOM} & \cite{SPM} & \cite{GFSDCF} & \cite{C-RPN} & \cite{UpdateNet} & \cite{DiMP} & \cite{DiMP} & \cite{UPDT} & \cite{ECO} & \cite{SiamRPN++} & \cite{DaSiamRPN}\\
\midrule
Precision   & $67.4$  & $\color{red}\text{70.0}$ & $64.8$  & $66.1$ & $56.6$  & $61.9$ & $62.5$  & $66.6$ & $\color{green}\text{68.7}$ & $55.7$  & $49.2$ & $\color{blue}\text{69.4}$  & $41.3$  \\
Norm. Prec. & $79.3$ & $\color{red}\text{81.2}$ & $77.1$  & $77.8$ & $71.8$  & $74.6$ & $75.2$  & $75.8$  & $\color{blue}\text{80.1}$ & $70.2$  & $61.8$ & $\color{green}\text{80.0}$  & $60.2$\\
Success   & $72.6$  & $\color{red}\text{74.9}$ & $70.3$  & $71.2$ & $60.9$  & $66.9$ & $67.7$  & $72.3$ & $\color{blue}\text{74.0}$ & $61.1$  & $55.4$ & $\color{green}\text{73.3}$  & $56.8$\\
\bottomrule
\end{tabular}
\end{table*}

\noindent \textbf{Experiment on LaSOT \cite{Lasot} dataset}: The LaSOT dataset is that consists of $1,400$ video sequences, with more than 3.5M image frames, and $280$ videos in the testing set. To validate the tracking accuracy, we conduct several experimental comparisons on LaSOT testing set in order to assess our proposed DIoUTrack alongside some state-of-the-art tracking methods, namely MDNet \cite{MDNet}, ECO \cite{ECO}, CFNet \cite{CFNet}, PTAV \cite{PTAV}, BACF \cite{BACF}, DSiam \cite{DSiam}, StructSiam \cite{StructSiam}, VITAL \cite{VITAL}, STRCF \cite{STRCF}, TRACA \cite{TRACA}, SiamRPN++ \cite{SiamRPN++}, ASRCF \cite{ASRCF}, GCT \cite{GCT}, ATOM \cite{ATOM}, DiMP18 \cite{DiMP}, DiMP50 \cite{DiMP}, UpdateNet \cite{UpdateNet}, ROAM \cite{ROAM}, SiamBAN \cite{SiamBAN}, SiamCAR  \cite{SiamCAR}, LTMU \cite{LTMU}, CLNet \cite{CLNet} and Ocean  \cite{Ocean}. Table \ref{tab:LaSOT} present the results of this comparison. Among these compared trackers, the DiMP50 \cite{DiMP} obtains the second best precision, normalized precision and success scores. In contrast, our DIoUTrack50 outperforms the DiMP50 \cite{DiMP} tracker on each performance metric item, which fully proves the effectiveness of our tracker.

\begin{table*}  [http]
\renewcommand\arraystretch{1.2}
\renewcommand\tabcolsep{3pt}
\scriptsize
\centering
\caption{Comparison results on LaSOT dataset. The top-3 scores are highlighted in {\color{red}\text{red}}, {\color{blue}\text{blue}} and {\color{green}\text{green}}, respectively.}
\label{tab:LaSOT}
\begin{tabular}{lcccc}\toprule
Trackers & Reference    & Precision scores & Normalized Precision scores & Success scores\\
\midrule
    MDNet \cite{MDNet} & CVPR2016        & $37.0$ & $37.3$ & $39.7$ \\
    ECO  \cite{ECO}  & CVPR2017          & $30.2$ & $30.1$ & $32.4$ \\
    CFNet  \cite{CFNet} & CVPR2017       & $26.3$ & $25.9$ & $27.5$ \\
    PTAV   \cite{PTAV}  & ICCV2017       & $24.5$ & $25.4$ & $25.0$ \\
    BACF   \cite{BACF}  & ICCV2017       & $23.6$ & $23.9$ & $25.9$ \\
    DSiam \cite{DSiam}  & ICCV2017       & $31.8$ & $32.2$ & $33.3$ \\
    StructSiam \cite{StructSiam} & ECCV2018 & $32.6$ & $33.3$ & $33.5$ \\
    VITAL  \cite{VITAL} & CVPR2018       & $36.2$ & $36.0$ & $39.0$ \\
    STRCF  \cite{STRCF} & CVPR2018       & $29.6$ & $29.8$ & $30.8$ \\
    TRACA  \cite{TRACA}  & CVPR2018      & $23.0$ & $22.7$ & $25.7$ \\
    SiamRPN++ \cite{SiamRPN++} & CVPR2019 & $46.7$ & $49.1$ & $49.6$ \\
    ASRCF  \cite{ASRCF} & CVPR2019       & $32.8$ & $33.1$ & $34.4$ \\
    GCT  \cite{GCT} & CVPR2019           & $32.8$ & $33.1$ & $34.4$ \\
    ATOM  \cite{ATOM}  & CVPR2019        & $47.9$ &$50.5$ & $51.4$ \\
    DiMP18  \cite{DiMP}  & ICCV2019      & $51.1$ & $54.1$ & $53.7$ \\
    DiMP50  \cite{DiMP}  & ICCV2019  & \color{blue}\text{53.8}  & \color{blue}\text{56.9} & \color{blue}\text{57.1}\\
    UpdateNet \cite{UpdateNet}  & ICCV2019 & 44.2     & 45.9  & 47.5  \\
    ROAM  \cite{ROAM}  & CVPR2020         &  35.8     & 36.8 &  39.0 \\
    SiamBAN  \cite{SiamBAN}  & CVPR2020      & 49.1   & 52.1 &  51.4 \\
    SiamCAR  \cite{SiamCAR}  & CVPR2020      & 48.1   & 51.0 & 50.7  \\
    LTMU  \cite{LTMU}  & CVPR2020         & 50.8      & 53.5 &  53.9 \\
    CLNet  \cite{CLNet}  & ECCV2020      &47.0        & 49.4 & 49.9  \\
    Ocean  \cite{Ocean}  & ECCV2020  & \color{green}\text{53.3}  & \color{green}\text{56.6} & \color{green}\text{56.0}\\
    DIoUTrack18   & Ours   & $51.6$ & $54.7$ & $54.7$ \\
    DIoUTrack50   & Ours   & $\color{red}\text{54.6}$ & $\color{red}\text{57.7}$ & $\color{red}\text{57.9}$ \\
\bottomrule
\end{tabular}
\end{table*}

\noindent \textbf{Experiment on GOT10k \cite{GOT10k} dataset}: This GOT10k test set includes $180$ video sequences for evaluation of tracking performance.
We conduct experimental comparisons on GOT10k test set to evaluate the tracking performance of our DIoUTrack relative to other state-of-the-art trackers, namely MDNet \cite{MDNet}, ECO \cite{ECO}, DSiam \cite{DSiam}, DAT \cite{DAT}, DeepSTRCF \cite{STRCF}, STRCF \cite{STRCF}, SASiamP \cite{SASiam}, SASiamR \cite{SASiam}, MemDTC \cite{MemDTC}, MetaSDNet \cite{MetaSDNet}, RT-MDNet \cite{RT-MDNet}, LDES \cite{LDES}, SiamDW \cite{SiamDW}, SPM \cite{SPM}, ATOM \cite{ATOM}, DiMP18 \cite{DiMP}, DiMP50 \cite{DiMP}, SiamCAR \cite{SiamCAR}, ROAM \cite{ROAM}, and  Ocean \cite{Ocean}. The comparison results are presented in Table \ref{tab:GOT10k}. The ATOM tracker obtains an average overlap score of $55.6\%$; however, our DIoUTrack18 with the same backbone network (ResNet18) achieves a $3.9\%$ performance improvement over the ATOM tracker, as well as faster tracking speed.  Meanwhile, our DIoUTrack50 achieves the best AO score and SR$_{0.50}$ score. Although our tracking speed is slightly lower than that of SPM \cite{SPM} tracker, in terms of tracking accuracy, our tracker obviously exceeds the SPM \cite{SPM} tracker in each indicator.

\begin{table*} [http]
\renewcommand\arraystretch{1.2}
\renewcommand\tabcolsep{3pt}
\scriptsize
\centering
\caption{Comparison results on GOT10k dataset. The top-3 scores are highlighted in {\color{red}\text{red}}, {\color{blue}\text{blue}} and {\color{green}\text{green}}, respectively.}
\label{tab:GOT10k}
\begin{tabular}{lccccc}
\toprule
Trackers & Reference  & Average Overlap (AO) & Success Rate$_{0.50}$ (SR$_{0.50}$) & Success Rate$_{0.75}$ (SR$_{0.75}$) & Speed \\
\midrule
    MDNet \cite{MDNet} & CVPR2016        & $35.2$ & $36.7$ & $13.7$ & $0.95$ \\
    ECO   \cite{ECO}   & CVPR2017        & $39.5$ & $40.7$ & $17.0$ & $2.21$\\
    DSiam \cite{DSiam}  & ICCV2017       & $41.7$ & $46.1$ & $14.9$ & $3.78$\\
    DAT    \cite{DAT}   & NIPS2018       & $41.1$ & $43.2$ & $14.5$ & $0.08$\\
    DeepSTRCF \cite{STRCF} & CVPR2018    & $44.9$ & $48.1$ & $16.9$ & $10.70$\\
    STRCF \cite{STRCF}  & CVPR2018       & $37.7$ & $38.7$ & $15.1$ & $3.06$\\
    SASiamP \cite{SASiam}  & CVPR2018    & $44.5$ & $49.1$ & $16.5$ & $25.40$\\
    SASiamR \cite{SASiam}  & CVPR2018    & $44.3$ & $49.2$ & $16.0$ & $5.13$\\
    MemDTC \cite{MemDTC} & ECCV2018      & $46.0$ & $52.3$ & $19.3$ & $0.35$\\
    MetaSDNet \cite{MetaSDNet} &ECCV2018 & $40.4$ & $42.3$ & $15.6$ & $0.53$\\
    RT-MDNet \cite{RT-MDNet} & ECCV2018  & $40.4$ & $42.4$ & $14.7$ & $7.85$\\
    LDES   \cite{LDES}     & AAAI2019    & $35.9$ & $36.8$ & $15.3$ & $1.23$\\
    SiamDW \cite{SiamDW} & CVPR2019      & $41.1$ & $45.6$ & $15.4$ & $12.00$\\
    SPM  \cite{SPM} & CVPR2019           & $51.3$ & $59.3$ & $35.9$ & $\color{red}\text{72.30}$\\
    ATOM  \cite{ATOM} & CVPR2019         & $55.6$ & $63.4$ & $40.2$  & $20.71$\\
    DiMP18  \cite{DiMP}  & ICCV2019      & $57.9$ & $67.2$ & $44.6$ & $34.05$\\
    DiMP50 \cite{DiMP}     & ICCV2019    & $\color{blue}\text{61.1}$ & $\color{green}\text{71.7}$ & $\color{red}\text{49.2}$   & $43.0$ \\
    SiamCAR \cite{SiamCAR} & CVPR2020    & $56.9$ & $67.0$ & $41.5$   & $\color{green}\text{52.30}$ \\
    ROAM \cite{ROAM}       & CVPR2020    & $43.6$ & $46.6$ & $16.4$   & $13.00$ \\
    Ocean \cite{Ocean}     & ECCV2020    & $\color{blue}\text{61.1}$ & $\color{blue}\text{72.1}$ & $\color{blue}\text{47.3}$   & $44.20$ \\
    DIoUTrack18    & Ours                & $59.5$ & $70.4$ & $44.0$ & $\color{blue}\text{53.46}$\\
    DIoUTrack50    & Ours                & $\color{red}\text{61.4}$ & $\color{red}\text{73.7}$ & $\color{blue}\text{47.3}$ & $44.15$\\
\bottomrule
\end{tabular}
\end{table*}

\noindent \textbf{Experiment on VOT2018 \cite{VOT2018} dataset}:
VOT2018 is containing $60$ test video sequences, and trackers  are measured using the expected average overlap (EAO), robustness and accuracy.
We make some comparisons of our tracker with several state-of-the-art trackers, namely ATOM \cite{ATOM}, DiMP18 \cite{DiMP}, DiMP50 \cite{DiMP}, PrDiMP18 \cite{PrDiMP}, PrDiMP50 \cite{PrDiMP} DaSiamRPN \cite{DaSiamRPN}, SiamRPN++ \cite{SiamRPN++}, UPDT \cite{UPDT} and Ocean \cite{Ocean} on this test set.
The comparison results are shown in Table \ref{tab:vot2018}. Our DIoUTrack18 has the best accuracy score compared to other trackers. Our DIoUTrack18 adopts the same backbone network as ATOM \cite{ATOM}, DiMP18 \cite{DiMP} and PrDiMP18 \cite{PrDiMP} trackers, and the EAO score and accuracy score are all higher than these trackers, which significantly indicates that our proposed method can bring more accurate tracking results.

\begin{table*}
\renewcommand\arraystretch{1.2}
\renewcommand\tabcolsep{2pt}
\scriptsize
\centering
\caption{Comparison results on VOT2018 dataset.}
\label{tab:vot2018}
\begin{tabular}{lcccccccccccc}
\toprule
\multirow{2}{*}{Trackers} & DIoUTrack18  & DIoUTrack50 & ATOM & DiMP18 & DiMP50 & PrDiMP18 & PrDiMP50 & DaSiamRPN & SiamRPN++ & UPDT & Ocean\\
& Ours & Ours & \cite{ATOM} & \cite{DiMP} & \cite{DiMP} & \cite{PrDiMP} & \cite{PrDiMP} & \cite{DaSiamRPN} & \cite{SiamRPN++} & \cite{UPDT} & \cite{Ocean}\\
\midrule
EAO ($\uparrow$)          & 0.435  & \color{blue}\text{0.444}  & 0.401 & 0.402 & 0.440 & 0.385 & \color{green}\text{0.442} & 0.383 & 0.414 & 0.378 &  \color{red}\text{0.467}\\
Robustness ($\downarrow$) & 0.185  & \color{red}\text{0.143}  & 0.204 & 0.182 &  \color{blue}\text{0.153} & 0.217 & \color{green}\text{0.165} & 0.276 & 0.234 & 0.184 & 0.169\\
Accuracy ($\uparrow$)     &  \color{red}\text{0.619}  & \color{green}\text{0.607}  & 0.590 & 0.594 & 0.597 & \color{green}\text{0.607} & \color{blue}\text{0.618} & 0.586 & 0.600 & 0.536 & 0.598\\
\bottomrule
\end{tabular}
\end{table*}

\noindent \textbf{Experiment on VOT2019 \cite{VOT2019} dataset}:
VOT2019 has the same data set size as VOT2018, and trackers are also evaluated using the expected average overlap (EAO), robustness and accuracy.
We make some comparisons of our tracker with several state-of-the-art trackers, namely ATOM \cite{ATOM}, DiMP18 \cite{DiMP}, DiMP50 \cite{DiMP}, PrDiMP18 \cite{PrDiMP}, PrDiMP50 \cite{PrDiMP}, TADT \cite{TADT}, SiamRPN++ \cite{SiamRPN++}, MemDTC \cite{MemDTC} and  Ocean \cite{Ocean} on this test set.
The comparison results are shown in Table \ref{tab:vot2019}.  Our DIoUTrack50 has the best EAO score compared to other trackers. Our DIoUTrack50 adopts the same backbone network as DiMP50 \cite{DiMP}, PrDiMP50 \cite{PrDiMP} and SiamRPN++ \cite{SiamRPN++} trackers, and the EAO score and accuracy score are all higher than these trackers, which also indicates that our proposed method can bring more accurate tracking results.

\begin{table*}
\renewcommand\arraystretch{1.2}
\renewcommand\tabcolsep{2pt}
\scriptsize
\centering
  \caption{Comparison results on VOT2019 dataset.}
  \label{tab:vot2019}
  \begin{tabular}{lcccccccccccc}
    \toprule
\multirow{2}{*}{Trackers} & DIoUTrack18  & DIoUTrack50 & ATOM & DiMP18 & DiMP50 & PrDiMP18 & PrDiMP50 & TADT & SiamRPN++ & MemDTC & Ocean\\
          & Ours & Ours & \cite{ATOM} & \cite{DiMP} & \cite{DiMP} & \cite{PrDiMP} & \cite{PrDiMP} & \cite{TADT} & \cite{SiamRPN++} & \cite{MemDTC} & \cite{Ocean}\\
    \midrule
EAO ($\uparrow$)          & $0.298$ & \color{red}\text{0.380}  & 0.299  & 0.318  &  \color{blue}\text{0.368}  & 0.314 & 0.268 & 0.207  & 0.285  & 0.228 & \color{green}\text{0.327}\\
Robustness ($\downarrow$) & 0.393 &  \color{red}\text{0.261}  & 0.411  & 0.369  & \color{blue}\text{0.278}  & \color{green}\text{0.355} &0.429 & 0.677  & 0.482 & 0.587 & 0.376 \\
Accuracy ($\uparrow$)     &  \color{blue}\text{0.609} & \color{blue}\text{0.609}  & 0.606  & 0.595  & 0.597  & \color{red}\text{0.611} & 0.572 & 0.516  & 0.599 & 0.485  & 0.590 \\
    \bottomrule
  \end{tabular}
\end{table*}

\begin{figure*} [!t]
\centering
\centerline{{\includegraphics[width=0.96\textwidth]{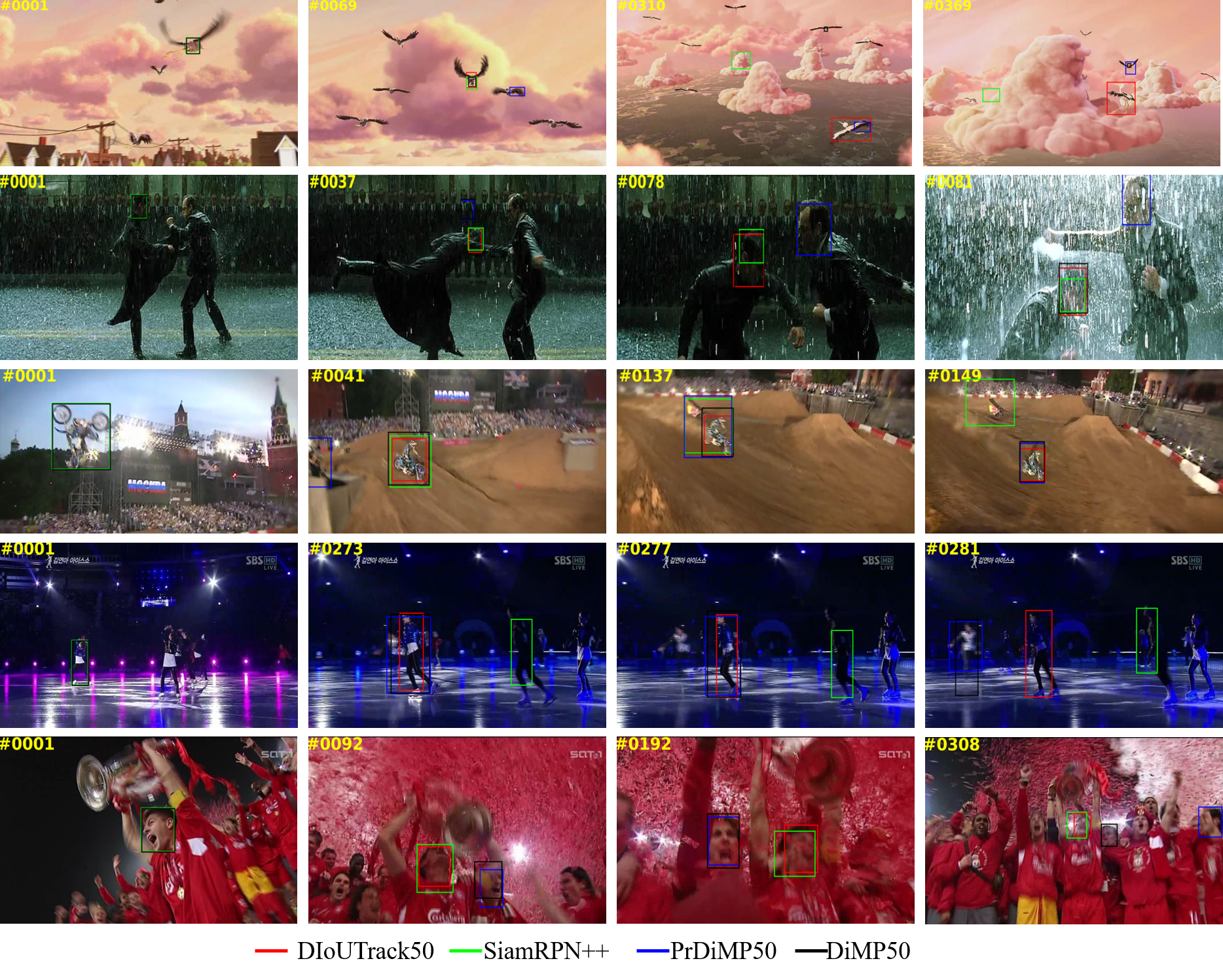}}}
\caption{Qualitative comparison (sequences from top to bottom are: bird1, matrix, motorrolling, skating1 and soccer selected from the OTB100 dataset).}
\label{fig:qua}
\end{figure*}

\subsection{Qualitative comparison}
To visually demonstrate tracking results, we present a qualitative comparison of our DIoUTrack50 to some state-of-the-art tracking methods, namely SiamRPN++ \cite{ SiamRPN++}, PrDiMP50 \cite{PrDiMP} and DiMP50 \cite{DiMP}.
All of these trackers utilize the same backbone network: ResNet50.
Figure \ref{fig:qua} presents these visual comparison results on several of the most challenging sequences selected from the OTB100 \cite{OTB100} dataset. For the DiMP50 \cite{DiMP} tracker, it interferes easily in the scenes of occlusion, fast motion, background cluster, and deformation ($e.g.$, bird1, and soccer). One explanation for this drawback is that it adopts the bounding-box regression model improved by an IoUNet, meaning that it is unable to locate the target accurately in some complex tracking scenes. By contrast, the proposed DIoUTrack50 adopts a distance-IoU network to improve the bounding-box regression model; this means that when the IoU score is constant, our model selects the candidate with the more accurate center position as the target. Meanwhile, the PrDiMP50 \cite{PrDiMP} tracker can not achieve ideal tracking results in illumination variation, deformation, scale variation and other aspects of tracking scenarios ($e.g.$, bird1, matrix and soccer). Moreover, the SiamRPN++ \cite{SiamRPN++} tracker readily interferes in the scenes of fast motion, scale variation, and deformation ($e.g.$,  motorrolling, skating1 and soccer); by contrast, our DIoU-based DIoUTrack50 obtains accurate tracking results on these testing video sequences.
In summary, compared with these state-of-the-art trackers, our proposed tracker produces more accurate boundary boxes and tracking results.

\section{Conclusions} \label{Con}
In this work, we propose an accurate bounding-box regression tracking method based on the distance-intersection-over-union (DIoU) loss. The proposed tracker comprises two components: an estimation component and a classification component. The former is trained offline in order to predict the DIoU overlap score between the target ground-truth and the predicted bounding box. Compared with the IoU loss, the adopted DIoU loss can make the prediction result closer to the real target in the training stage, which can predict the target boundary box more accurately in the tracking process. While the classification component is trained online using the Conjugate-Gradient-based method, resulting in a fast-tracking speed. Extensive experimental results on seven challenging benchmarks show that our proposed method obtains competitive tracking results compared with state-of-the-art trackers.
Our future work will focus on how to better use large-scale unlabeled data to train the CNN model of the tracker. We hope to apply an unsupervised domain adaptation method for the tracker to achieve this goal.

\section*{Acknowledgement}
This research was supported by the National Natural Science Foundation of China (Grant No. 62172126), by the Special Research project on COVID-19 Prevention and Control of Guangdong Province (Grant No. 2020KZDZDX1227), by the Shenzhen Research Council (Grant No. JCYJ20210324120202006).  Di Yuan was supported by a scholarship from China Scholarship Council. Dr Xiaojun Chang was partially supported by Australian Research Council (ARC) Discovery Early Career Researcher Award (DECRA) under grant no. DE190100626.

\section*{References}

\bibliography{mybibfile}

\end{document}